\documentclass[letterpaper]{article} 
\usepackage{aaai24}  
\usepackage{times}  
\usepackage{helvet}  
\usepackage{courier}  
\usepackage[hyphens]{url}  
\usepackage{graphicx} 
\usepackage{natbib}  
\usepackage{caption} 
\usepackage{algorithm}
\usepackage{algorithmic}

\usepackage{epsfig}
\usepackage{amsmath}
\usepackage{amssymb}
\usepackage{float}
\usepackage{subfig}
\usepackage{booktabs}
\usepackage{colortbl}
\usepackage{color}
\usepackage{pifont}
\usepackage{bbding}
\usepackage{multirow}
\usepackage{epstopdf}
\definecolor{tabcolor}{rgb}{.0,.0,.0}
\usepackage{nicematrix}
\usepackage{makecell}
\usepackage{bm}

\usepackage{newfloat}
\usepackage{listings}
\DeclareCaptionStyle{ruled}{labelfont=normalfont,labelsep=colon,strut=off} 
\floatstyle{ruled}
\newfloat{listing}{tb}{lst}{}
\floatname{listing}{Listing}
\title{HybridGait: A Benchmark for Spatial-Temporal Cloth-Changing Gait Recognition with Hybrid Explorations}
\author{
    Yilan Dong\textsuperscript{\rm 1},
    Chunlin Yu\textsuperscript{\rm 1},
    Ruiyang Ha\textsuperscript{\rm 1},\\
    Ye Shi\textsuperscript{\rm 1},
    Yuexin Ma\textsuperscript{\rm 1},
    Lan Xu\textsuperscript{\rm 1},
    Yanwei Fu\textsuperscript{\rm 2},
    Jingya Wang\textsuperscript{\rm 1}\thanks{Corresponding author.} 
}
\affiliations{

    \textsuperscript{\rm 1}ShanghaiTech University
    \textsuperscript{\rm 2}Fudan University\\
    yilandong@outlook.com; yanweifu@fudan.edu.cn; \{yuchl, hary2022, shiye, mayuexin, xulan1, wangjingya\}@shanghaitech.edu.cn;
    
}

\usepackage{bibentry}

\begin{document}

\maketitle

\begin{abstract}
Existing gait recognition benchmarks mostly include minor clothing variations in the laboratory environments, but lack persistent changes in appearance over time and space. In this paper, we propose the first in-the-wild benchmark CCGait for cloth-changing gait recognition, which incorporates diverse clothing changes, indoor and outdoor scenes, and multi-modal statistics over 92 days. To further address the coupling effect of clothing and viewpoint variations, we propose a hybrid approach HybridGait that exploits both temporal dynamics and the projected 2D information of 3D human meshes. Specifically, we introduce a Canonical Alignment Spatial-Temporal Transformer (CA-STT) module to encode human joint position-aware features, and fully exploit 3D dense priors via a Silhouette-guided Deformation with 3D-2D Appearance Projection (SilD) strategy. Our contributions are twofold: we provide a challenging benchmark CCGait that captures realistic appearance changes across an expanded  and space, and we propose a hybrid framework HybridGait that outperforms prior works on CCGait and Gait3D benchmarks.
Our project page is available at 
\textit{\color{magenta}{https://github.com/HCVLab/HybridGait}}.
\end{abstract}
\section{Introduction}

Gait recognition is an intriguing biometric task that aims to identify individuals based on their unique walking patterns. Unlike traditional biometric identifiers such as faces, fingerprints, and irises, gait information is a non-cooperative identifier as well as a non-invasive differentiator, making it a effective and secure option. Consequently, gait recognition has garnered significant interest in the research community, with potential applications in security, surveillance, crime analysis and forensic search.

Extracting reliable gait features while mitigating distractions like viewpoint, and pose variations has long been a challenge in the gait community. To date, the role of clothing and accessories has also gaining increasing attention towards more practical scenarios. 
More recently, the long-term vision \cite{DeepChange} presents a new challenge for cloth-changing person re-identification, involving non-stationary cloth-changes over a long time scale. Unfortunately, recent advances in gait recognition \cite{CASIA-B,OU-ISIR-Cloth} typically have a short-term in-the-lab assumption that only a fixed gallery of clothes is imposed to mimic such variations at a fixed location, thereby disregarding the natural changes in appearance over time and space. The very recently proposed benchmarks \cite{GRAW, Gait3D} break the previous laboratory assumption by introducing an in-the-wild setup, however, they not only overlook the variations in clothing and accessories but also fail to account for changes over time.


To address the aforementioned issues, we present the first in-the-wild cloth-changing gait recognition benchmark, CCGait, which captures appearance changes over time and space. Our dataset can facilitate the development of more robust and resilient gait recognition systems that can perform effectively in real-world situations especially for cloth-changing scenarios. Critically,
our proposed dataset possesses unique features that set it apart from pre-existing datasets \cite{Gait3D, GREW, OU-MVLP, CASIA-B}. Firstly, the raw videos were captured over a 92-day period, providing a diverse range of cloth changes occurring at various frequencies without human intervention. Secondly, our dataset includes footage from both indoor and outdoor surveillance systems, capturing individuals walking along different routes and at varying speeds. Lastly, the CCGait tool provides a range of multi-modal statistics, such as 2D silhouettes, 2D/3D keypoints, and 3D human meshes. These features make our dataset a valuable resource for researchers in the field, enabling them to analyze and evaluate gait recognition algorithms under different conditions and scenarios.

We look at the problem of cloth-changing gait recognition under in-the-wild scenarios. Previous works utilize 2D silhouette information for robust extraction of gait features, while more recently 3D human meshes have been exploited to leverage the invariant properties against clothing and viewpoint variations \cite{Gait3D, han2022licamgait}. However, to enable the integration of human meshes with silhouettes, existing approaches typically adopt a simple and straightforward fusion strategy, which may be inadequate in bridging the modality gap and extracting mutual information from two modalities. On this ground, a more careful design is needed to unify the two sharply different representation structures, and make full use of 3D meshes.

To implement this vision, we propose a SMPL-aided hybrid network comprising three branches, where a temporal branch and a projection branch are devised to assist the appearance branch. In the temporal branch, we develop the \textit{Canonical Alignment Spatial-Temporal Transformer} (\textbf{CA-STT}) which serves a dual purpose: firstly, it captures the cloth-irrelevant intrinsic temporal dynamics of 3D human mesh; secondly, it constructs a canonical space to bridge the gap between the 3D human mesh and the 2D rest pose by explicitly aligning the kinematic and regular grid features. In the projection branch, we first project 3D meshes to 2D silhouettes at a specific view. The projection is executed prior to training, ensuring no additional training or inference time. While the 3D SMPL projected silhouettes can eliminate large viewpoint variations, they may lack crucial appearance details. Hence, we propose a deformable alignment strategy named \textit{Silhouette-guided Deformation} (\textbf{SilD}), which enriches the semantic content of projection appearance by leveraging original silhouettes as intermediary guidance.


We summarize our \textbf{contributions} as follows:
\begin{itemize}
\item Firstly, we introduce a benchmark CCGait tailored for in-the-wild clothing-change gait recognition. This dataset is designed to facilitate pragmatic research in gait recognition, specifically addressing challenges associated with appearance changes over time and space.
\end{itemize}

\begin{itemize}
\item Secondly, we propose a novel hybrid framework HybridGait that leverages both the projected 2D appearance and temporal dynamics of 3D human mesh. Our framework includes a Canonical Alignment Spatial-Temporal Transformer to capture the clothes-irrelevant dynamics of gait in the temporal branch, and a Silhouette-guided Deformation to address variations in viewpoint in the projection branch.
\end{itemize}

\begin{itemize}
\item Finally, our experimental results demonstrate that HybridGait obtains consistent improvements for gait recognition on both CCGait and Gait3D benchmarks. 
\end{itemize}

\begin{table*}[htbp]
\begin{center}
\resizebox{\textwidth}{!}{
\begin{NiceTabular}{l|c|c|c|c|c|c|c|c}
\arrayrulecolor{tabcolor}
\toprule[2pt]
Dataset & IDs & Tracklets & Views & Data Tpyes & Environment & Time(Day) & Cloth Change & No Human Intervention \\
\midrule
\midrule
CCVID & 226 & 2,856 & 1 & RGB & Outdoor & - & \Checkmark & \XSolidBrush \\
CCPG & 200 & 16,566 & 10 & RGB, Sil. & Indoor \& Outdoor & - & \Checkmark & \XSolidBrush\\
\midrule

CASIA-B & 124 & 13,640 & 11 & RGB, Sil. & Indoor & - & \Checkmark & \XSolidBrush \\ 
OU-ISIR Cloth & 68 & 2,764 & 1 & Sil. & Indoor & - & \Checkmark & \XSolidBrush  \\
OU-LP Bag & 62,528 & 187,584 & 1 & Sil. & Indoor & - & \Checkmark & \XSolidBrush  \\
OU-MVLP & 10,307 & 288,596 & 14 & Sil. & Indoor & - & \XSolidBrush & \XSolidBrush \\
FVG & 226 & 2,856 & 1 & RGB & Outdoor & - & \Checkmark & \XSolidBrush  \\
GREW & 26,345 & 128,671 & 882 & Sil., Pose, Flow & Outdoor & 1  & \Checkmark & \Checkmark   \\
Gait3D & 4,000 & 25,309 & 39 & Sil., Pose, Mesh & Indoor & 7  & \XSolidBrush  & \Checkmark  \\
\midrule
CCGait & 1,495 & 4,824 & 9 & Sil., Pose, Mesh  & Indoor \& Outdoor & 92 & \Checkmark  & \Checkmark \\
\bottomrule[2pt]
\end{NiceTabular}
}
\end{center}
\caption{Comparison of publicly available datasets for video-based cloth-changing person re-identification and gait recognition.}
\label{tab1}
\end{table*}

\section{Related Work}

\noindent {\bf Gait Recognition}. Currently, the research lines of gait recognition can be roughly grouped into two categories: appearance-based methods and model-based methods. 
\textit{Appearance-based methods} \cite{GEINet,GaitSet,GLN,GaitPart,3DLocal,GaitGL,MT3D,MTSGait,CSTL} learn gait representations directly from binary visual cues of silhouette sequences. While Chao \textit{et al.} \cite{GaitSet} and Hou \textit{et al.} \cite{GLN} utilize unordered silhouette sequences for gait recognition,  recent works focus on the temporal dynamics with ordered silhouettes as input, either using 1D CNN to capture the temporal information \cite{GaitPart} or extracting spatial-temporal information via 3D convolutions \cite{MT3D, CSTL, 3DLocal}. Although appearance-based methods are concise and effective, they hold a strong assumption that appearance statistics of individuals only have moderate changes, which is not suitable for cloth-changing gait recognition. \textit{Model-based methods} \cite{AutoMated, Recognition, model-base, GaitGraph, model} aim at modeling
the structure of the human body from pose information, where Liao \textit{et al.} \cite{model} use 3D keypoints as human prior knowledge, while Teep \textit{et al.} \cite{GaitGraph} employ 2D skeletons to represent walking patterns. More recently, Zheng \textit{et al.} \cite{Gait3D} make the first attempt to explore 3D human mesh in gait recognition, which shows great potential in reducing the effect of long-term perturbations. However, they ignore the temporal dynamics and the projected appearance information within the 3D human meshes and the naive fusion approach does not maximize the utilization of 3D mesh information. Zhu \textit{et al.}\cite{HBS} directly extract 3D mesh information from silhouette to mitigate view and clothing disruptions. However, the accuracy of the 3D mesh information heavily rely on the original silhouette's quality, potentially causing challenges in in-the-wild scenarios. In stark contrast, our approach aims to characterize the temporal aand representation of the SMPL model and explicitly handle extreme viewpoint variations.

 {\bf Gait Recognition dataset}.
Current public datasets can be classified into two categories: in-the-lab and in-the-wild. Previous works mainly conduct experiments on two types of popular laboratory benchmarks such as CASIA series \cite{CASIA-A, CASIA-B, CASIA-C}) and OU-ISIR series \cite{OU-ISIR-Cloth, OU-ISIR-MV, OU-ISLR-LP, OU-ISIR-Speed, OU-MVLP, OU-LP-Bag}). While a few existing benchmark efforts such as CASIA-B \cite{CASIA-B}) and OU-ISIR Cloth \cite{OU-ISIR-Cloth} have initiated the research on cloth-changing gait recognition, they focus more on laboratory moderate cloth variations in constrained environments. Recently, several in-the-wild datasets have been built to fulfill the challenging real-world demands. GREW \cite{GREW} is collected by hundreds of cameras set in public places only within a single day, resulting in restricted clothing variations. Gait3D \cite{Gait3D} builds the first in-the-wild gait recognition dataset with 3D mesh modalities, enabling the community to explore dense 3D representations for the gait. However, Gait3D is collected in a location where individuals are less likely to repeat their presence frequently. As a result, it doesn't fulfill the criteria for real-world clothing change scenarios. To address this gap, we have constructed a new dataset that specifically focuses on clothing changes for gait recognition in the wild.

\section{The CCGait Dataset}
In order to facilitate research on gait recognition, we introduce a new Cloth-Changing gait recognition dataset called CCGait. This dataset offers several distinct features compared to existing datasets, which are listed in Table \ref{tab1}. Specifically, the CCGait dataset was collected over a period of 92 days, during which time we captured data on individuals exhibiting a wide range of attire and appearances. This diverse collection of data is particularly useful for developing and testing gait recognition algorithms that are robust to variations in clothing, footwear, and other physical attributes.
In addition to its temporal diversity, the CCGait dataset offers spatial diversity as well. Specifically, the dataset includes footage captured from both indoor and outdoor surveillance systems, providing a variety of environments in which people walk at varying speeds and along different routes. This variability in walking conditions is particularly valuable for researchers seeking to develop gait recognition algorithms that can operate effectively in real-world settings.
Finally, the CCGait dataset includes a range of multi-modal statistics that researchers can use to develop and test their algorithms. Specifically, the dataset provides 2D silhouettes, 2D/3D keypoints, and 3D human meshes, allowing researchers to evaluate the performance of their algorithms across multiple modalities. This multi-modal approach can help to improve the robustness and reliability of gait recognition algorithms in a variety of settings.

\begin{figure*}[t]
\begin{center}
   \includegraphics[width=1\linewidth]{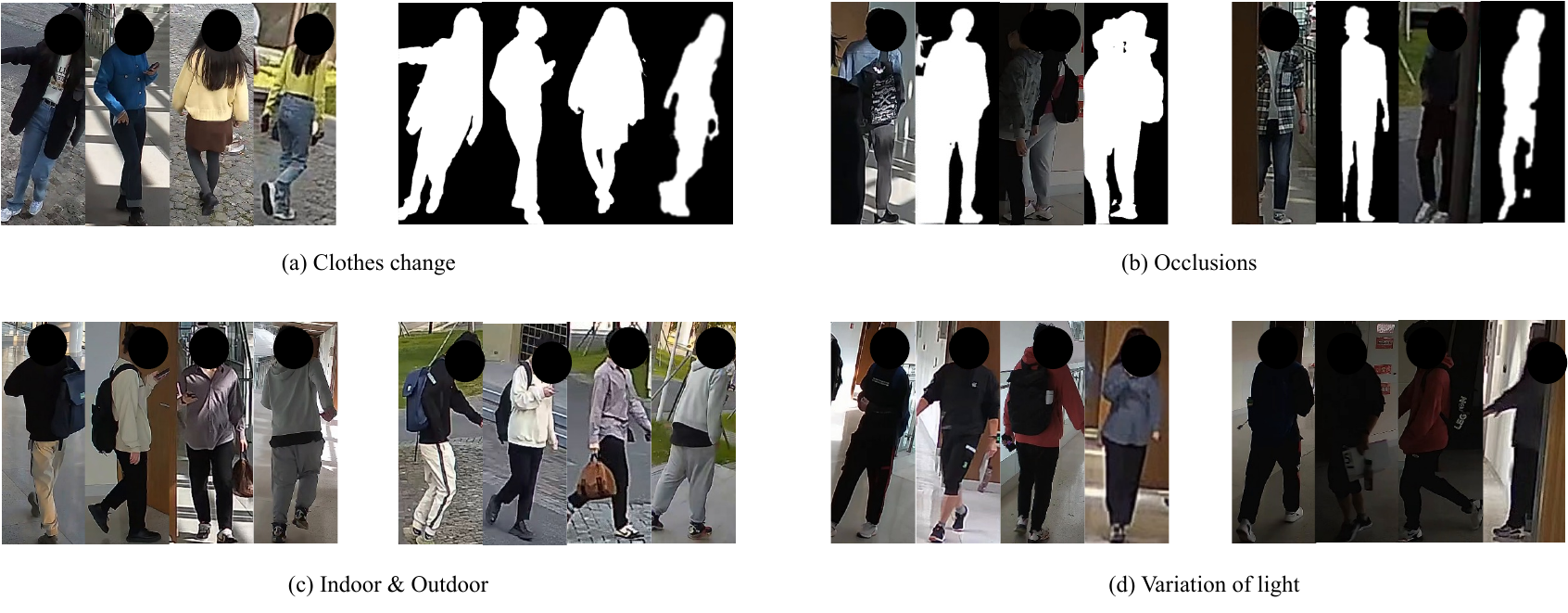}
\end{center}
   \caption{Exemplary challenge frames extracted from the CCGait dataset are presented in this paper. Figure (a) displays the long-ter cloth changes observed for the same individual. In Figure (b), we showcase occlusions caused by both persons and objects. Figure (c) showcases diverse indoor and outdoor scenarios, while Figure (d) demonstrates variations in lighting conditions.}
   \label{fig:challengefordataset}
\end{figure*}

\noindent \textbf{Data Collection.} We collected raw video footage from nine cameras located both outside and inside the building. As the locations we collected data from are places frequently visited by pedestrians on a daily basis, this greatly increases the probability of capturing the same individuals across a long period of time, even when they wear different clothing. The footage was captured at a resolution of 1,920 × 1,080 and at a frame rate of 25 FPS. 
For data collection, we chose to collect data during the peak pedestrian hours for 6 hours per day, 4 days per week, and collected data for over three-month period, totaling 2208 hours.
Except for clothes change conditions, the CCGait dataset also involves occlusions by humans and objects. Still, it contains both indoor and outdoor sequences, as well as the sequences collected under various light conditions. All the above attributes contribute to the challenge of the CCGait dataset.
Some examples of challenge cases can be seen in Figure \ref{fig:challengefordataset}.
The collection of the dataset has passed ethical review by the university. The recording has been authorized by the building administration and all persons involved have been notified to collect data only for research purposes. Furthermore, in order to protect privacy, we will not release any RGB images.

\noindent \textbf{Data Preprocessing and Annotation.} Prior to annotation, we utilized FairMOT \cite{FairMOT}, which had been fine-tuned to our dataset, to perform person tracking. We then engaged annotators to address any potential mistracking issues in the algorithm and to ensure that each sequence corresponded to only one person. Ultimately, we manually annotated a total of 4,824 sequences, resulting in 1,495 unique identities.

\noindent \textbf{Generation of Gait Representations.} 
The CCGait Dataset provides a comprehensive set of gait representations that include 3D SMPL parameters, 3D pose, 2D pose, and silhouette. 
Thanks to the rapid advancement of 3D pose and shape methods \cite{ROMP,zhang2023ikol,li2023niki}, we utilized ROMP \cite{ROMP}, which allowed us to obtain accurate 3D models of the human body.
For accurate estimation of body joints, we used PoseNet \cite{PoseNet} for 2D and 3D poses. This allowed us to obtain accurate estimates of body joint locations, which can be used to analyze human motion and detect abnormalities.
To obtain a 2D silhouette, we employed HRNet-segmentation \cite{HRNet-Segmentation}, which is a state-of-the-art method for semantic segmentation. 
To maintain accuracy, we kept the original resolution and aspect ratio of the frame during the generation. This helped to avoid any distortion or loss of information during the generation process. Some examples of multi-modal gait representations in our dataset can be found in Supplementary Material.

\noindent \textbf{Data Statistics .} 
The CCGait dataset is a collection of 4,824 sequences from 1,495 different individuals. 
The CCGait dataset is divided into train/test subsets with 1148/347 IDs, respectively. For the test set, we further randomly select one or two sequences from each ID to build the query set with 356 sequences, while the rest of the sequences become the gallery set with 727 sequences.
More statistical details can be found in Supplementary Material.

\begin{figure*}[htbp]
\begin{center}
   \includegraphics[width=1.0\linewidth]{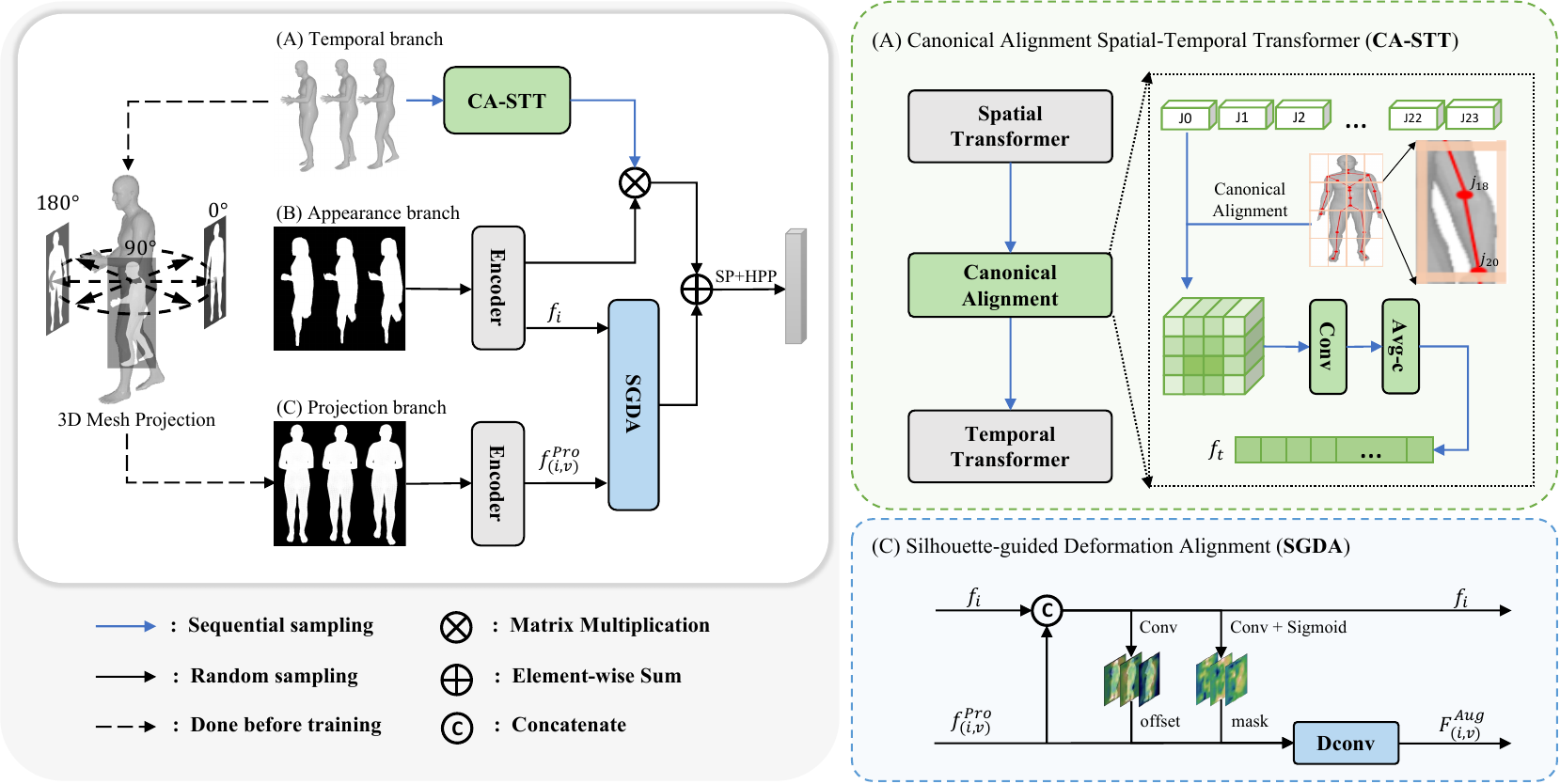}
\end{center}
   \caption{Our proposed framework comprises three key components. Component (B) is the basic appearance branch, responsible for extracting body representations from silhouettes, albeit susceptible to contextual perturbations. Component (A) is the temporal branch, which includes a 3D dynamic component, specifically the CA-STT model, to capture temporal information. Finally, component (C) is the Projection branch, which utilizes projected silhouettes from SMPL models as 3D-2D projection silhouettes to enhance the characterization of body representations.
}
\label{mainImg}
\end{figure*}

\section{Method}

\noindent\textbf{Method Overview.} Taking advantage of both the model-based methods and appearance-based methods, we consider a hybrid approach that jointly takes the 3D SMPL models and silhouettes as our input. Concretely, our framework is equipped with three branches. (1) The basic \textit{appearance branch} simply extracts body representations from silhouettes, which are susceptible to context perturbations. (2) The \textit{temporal branch} takes the pose statistics of SMPL models as input and aims to obtain temporal representations. (3) The \textit{projection branch} takes the projected silhouettes from SMPL models as input to further help characterize the body representations. The overall architecture of our method is shown in Figure \ref{mainImg}.

\noindent\textbf{Appearance Branch.} As the baseline of our approach, the appearance branch is adopted from \cite{Gait3D}, which directly takes silhouette sequences as input, denoted as $\{{\bm{x}_i}\}_{i=1}^N$, where $\bm{x}_i\in\mathbb{R}^{H\times W}$ represents the silhouette inputs, $N$ is the sequence length. We denote the extracted shape embeddings of $i$-th frame from the appearance branch as $\bm{F}=\{\bm{f}_i\}_{i=1}^N$. In the following sections, we first introduce our elaborately-designed components: the Canonical Alignment Spatial-Temporal Transformer (CA-STT) in the temporal branch 
, and the Silhouette-guided Deformation Alignment in the projection branch
Then, we introduce our training objectives and fusion strategy
.

\subsection{Canonical Alignment Spatial-Temporal Transformer} 
\label{CASTT}

 Inspired by \cite{3dhuman}, we establish a spatial-temporal transformer framework to extract pose and motion characteristics. On top of that, a rest pose canonical space is introduced to align the model features with the appearance features in a semantically consistent way.

\noindent {\bf Spatial-Temporal Transformer.} The goal of the spatial-temporal transformer is to learn the pose information across the frame scale and the motion statistics across the time scale. In the spatial transformer, the joint embeddings are first encoded into high-dimensional features, which are then fed into stacked spatial transformer layers. The output of the spatial transformer is denoted as: $\bm{F}^t_{sp} = \{ S_i^t\}_{i=1}^N$. For the $i$-th frame, the embedding of each joint is represented as: $\bm{S}_i^{t} = \{ \bm{f}_{ij}^t\}_{j=1}^{N_j}$, where $\bm{f}_{ij}^t\in \mathbb{R}^{C}$ refers to the $j$-th joint feature in the $i$-th frame, $N_j$ is the SMPL joint number and $C$ is the embedding dimension. Then, the joint features in each frame are averaged to obtain the frame representation. After that, frame embeddings are fed into temporal transformer layers to obtain the motion embedding $\bm{F}^t$.

\noindent {\bf Rest Pose Canonical Alignment.} However, the model and shape embeddings is not semantic-consistent within the feature space. Inspired by previous works on 3D reconstruction, which utilize pixel alignment in a canonical space \cite{arch++}, 
we first introduce a Rest Pose Canonical space, where each human joint is assigned a coordinate within a fixed-sized 2D image. Subsequently, we present a novel approach termed Rest Pose Canonical Alignment, which involves scaling these coordinates to align with the target size image.

Given the $i$-th frame feature $\bm{S}_i^t$ generated by the spatial transformer, our goal is to transform $\bm{S}_i^t\in\mathbb{R}^{d\times N_j}$ into a target feature map $\bm{S}_i^{t'} \in \mathbb{R}^{d\times (h\times w)} $, with $(h, w)$ denoting the predetermined resolution of the feature map. In our approach, we set $(h \times w)$ to the feature size of the appearance branch. However, a direct mapping lacks effective supervision. Observing that the regular grid structure retains the relative positioning of human joints, therefore, we convert this challenge into a task of finding the most pertinent human joint points for each pixel in the target space.

We first formulate the coordinate of the target patch regions as $\mathcal{R}=\{(h_r, w_r)\}_{r=1}^{N_r}$, where $N_r = h\times w$ is the number of target patch regions. Then the original rest pose canonical coordinate is defined as $\mathcal{C}=\{(h_j, w_j)\}_{j=1}^{N_j}$, where $h_j\in \{0, \cdots, H\}$ and $w_j\in \{0, \cdots, W\}$, $H$ and $W$ refers to the maximum values for the ordering of keypoints in the vertical and horizontal directions, respectively. After that, we can generate the proportionally scaled target canonical coordinate, $\mathcal{C'}=\{(\frac{h_j\times h}{H}, \frac{w_j\times w}{W})\}_{j=1}^{N_j}$. Then, the rest pose canonical alignment can be formulated as:
\begin{equation}
\mathcal{A} = \left\{(r, \left\{ \omega_{j, r}\right\}_{j=1}^{N_J} ): r \in \mathcal{R},\;\omega_{j, r}\in \{0, 1\} \right\}, 
\end{equation}
where $\{\omega_{j, r}\}_{j=1}^{N_J}$ can be interpreted as: for each patch region $r$, $\{\omega_{j, r}\}_{j=1}^{N_J}$ filters the irrelevant joints by enforcing the corresponding $\omega_{j, s}=0$ and retaining semantic-relevant joints.

We employ k-nearest neighbors to retrieve the k closest joints to the coordinates of the region $r$:
\begin{equation} \label{eq2}
    \{\omega_{j,r}\}_{j=1}^{N_j} = \text{KNN}(\mathcal{R}_r, \mathcal{C'}, k)
\end{equation}
Where $\text{KNN}$ returns 1 if keypoint $j$ is within the $k$-th closest joints, otherwise $\text{KNN}$ returns 0.

To further obtain the semantic content of patch region $r$ and subsequently achieve canonical alignments, we aggregate the related joint features using the obtained $\omega_{j, r}$. Formally, for each patch region $r\in \mathcal{R}$, we have:
\begin{equation}
(\bm{S}_i^{t'})^{(r)} =\text{Avg}(\sum_{j=1}^{N_j}\omega_{j,r}\bm{f}_{ij}^t)
\end{equation}
Then, the transformed feature $\bm{F}_{sp}^{t'}=\{\bm{S}_i^{t'}\}_{i=1}^{N_j}$ is passed through a modulate block:
\begin{equation}
\bm{F}^{t'} =\text{IF}(\text{Avgpool\_c}(\text{Conv}(\bm{F}_{sp}^{t'}))),
\end{equation}
where $\text{Avgpool\_c}$ refers to average pooling function along channel dimension, $\text{IF}\left(\cdot \right)$ refers to inverse flatten operation.

Notably, although the joints position is explicitly fixed, pose information is implicitly encoded by the spatial transformer, and then aligned to the 2D canonical space. Therefore, the temporal transformer can still learn the dynamic changes across time scales. 

\subsection{Silhouette-guided Deformable Alignment}

\label{SD3D}

To further enhance the shape information against viewpoint variations, we project the 3D SMPL model to uniform-viewed 2D silhouettes for all identities by \cite{mmhuman3d}. 
The projected $i$-th silhouette is denoted as $x^{Pro}_{(i,v)}$, $v$ is a specific viewpoint.
Then, the projected $i$-th silhouettes features, $\bm{f}_{(i,v)}^{Pro}$, are encoded with a feature extractor.

However, the direct pixel-wise fusion can lead to the misalignment of body parts between distorted silhouettes and viewpoint-consistent projected silhouettes. To tackle the aforementioned concerns, we propose utilizing the original silhouette embedding $f_i$ as a reference to guide the projected silhouette embedding $\bm{f}_{(i,v)}^{Pro}$ through the deformable convolution\cite{DefConv}. This involves formulating the corresponding deformation offset and modulating mask as follows:
\begin{equation}
\bm{\Phi} = \text{Conv\_o}([\bm{f}_i, \bm{f}_{(i,v)}^{Pro}]),
\end{equation}
\begin{equation}
\bm{m} = \text{Sigmoid}(\text{Conv\_m}([\bm{f}_i, \bm{f}_{(i,v)}^{Pro}])),
\end{equation}
where $\bm{\Phi}=\left\{ \Delta p_k | k = 1, ..., K \right\}$, $K$ is the size of convolution kernel, $[\cdot]$ is the concatenation operation. Then a semantic-enriched projected silhouette can be aligned to the reference image:
\begin{equation}
\bm{f}_{(i,v)}^{Pro'}(p) = \sum_{p_k\in K}\omega(p_k)\cdot \bm{f}_{(i,v)}^{Aug}(p+p_k+\Delta p_k)\cdot \Delta m_k,
\end{equation}
where $\bm{f}_{(i,v)}^{Pro'}$ is the output feature for the $i$-th frame, $p$ is a single pixel in the feature map, $p_k$ is the $k$-th regular offset of convolution kernal, $\Delta p_k$ and $\Delta m_k$ is the $k$-th learned offset and modulation scalar at location $p+p_k$.
In this way, the adaptively learned offset will capture semantic correlation cues with the assistance of the original silhouette in a pixel-level alignment, and the output feature is $\bm{F}^{Pro'}=\{\bm{f}_{(i,v)}^{Pro'}\}_{i=1}^N$.

\subsection{Training Objective}
\label{TO}

So far, with $\bm{F}$, $\bm{F}^{Pro'}$ and $\bm{F}^{t'}$, we produce the final embedding as following: 
\begin{equation}
    \hat{\bm{F}} = (\bm{F} \times \bm{F}^{t'}) + \bm{F}^{Pro'},
\end{equation}
where $\times$ is matrix multiplication. Finally, Set Pooling(SP) and Horizontal Pyramid Pooling(HPP)\cite{GaitSet} are adopted to obtain the final feature vector.

In the training stage, our three-branch framework is trained in an end-to-end manner, and a combined loss is adopted. The combined loss is defined as
\begin{equation} \label{eq6}
\mathcal{L} = \alpha\mathcal{L}_{tri} + \beta\mathcal{L}_{ce}, 
\end{equation}
where $\mathcal{L}_{tri}$ is the triplet loss, $\mathcal{L}_{ce}$ is the cross-entropy loss, $\alpha$ and $\beta$ are weighting hyperparameters.


\section{Experiments}

\noindent {\bf Datasets and Evaluation Protocol.}
In addition to our proposed CCGait benchmark, we conducted a comparison with the Gait3D dataset \cite{Gait3D}, which is currently the largest gait recognition dataset with 3D representations. To ensure consistency with previous gait recognition datasets \cite{OU-ISLR-LP,OU-ISIR-Cloth,OU-ISIR-MV,OU-ISIR-Speed,OU-LP-Bag,OU-MVLP,OU-MVLP-Pose,OUMVLP-Mesh,CASIA-A,CASIA-B,CASIA-C}, we used the same evaluation protocol, which involves open-set instance retrieval. Specifically, we measured the similarity between a given query sequence and all sequences in the gallery set and reported the accuracy using the average Rank-1 and Rank-5 identification rates across all query sequences. To account for the retrieval of multiple instances and difficult samples, we adopted two additional evaluation metrics: mean Average Precision (mAP) and mean Inverse Negative Penalty (mINP). These metrics provide a comprehensive evaluation of the performance of our proposed approach in comparison to the state-of-the-art methods.

\noindent{\bf Implementation Details.}
During the training process, we used the same configuration to train all models. The batch size is set as 32 × 4 × 30, where 32 represents the number of IDs, 4 for the training sequences per ID, and 30 denotes the sequence length. The models were trained for 1200 epochs using an initial learning rate (LR)=1e-3 and the LR was reduced by a factor of 0.1 at the 200-th and 600-th epochs. We use the loss in Equ. \ref{eq6} for training. The hyper-parameters in Equ. \ref{eq6} are set as $\alpha$=1.0 and $\beta$=0.1. And the hyper-parameters $k$ in Equ. \ref{eq2} is set to 7. We used the Adam optimizer \cite{Adam} and set the weight decay as 5e-4. The spatial-temporal transformer, encoder structure and more details can be found in Supplementary Material.

\subsection{Results and Analysis}
We present a comprehensive comparison of nine state-of-the-art (SOTA) models for gait recognition. Specifically, we evaluate the performance of these models in three categories. In the model-based method category, we analyze GaitGraph \cite{GaitGraph}, which utilizes a 2D skeleton as a graph and inputs it into a Graph Convolution Network. In the appearance-based method category, we examine GEINet \cite{GEINet}, GaitSet \cite{GaitSet}, GaitGL \cite{GaitGL}, GaitPart \cite{GaitPart}, CSTL \cite{CSTL}, MTSGait \cite{MTSGait} and GaitGCI\cite{GaitGCI}. These models use Convolutional Neural Networks (CNNs) to learn features from various sources, including GEIs, silhouettes, and gait sequences. GaitPart divides a silhouette image into fixed parts to learn micro-motion features. CSTL learns both long-term and short-term motion for gait recognition, and MTSGait learns spatial features and multi-scale temporal dynamics. Furthermore, we examine SMPLGait \cite{Gait3D}, which belongs to neither the model-based nor the appearance-based method category. This model utilizes SMPL \cite{smpl} in conjunction with a 2D silhouette to learn gait information.

The evaluation results of our CCGait Dataset are presented in Table \ref{tab3}. The findings indicate that: \textbf{(1)} the model-based method, GaitGraph, achieves only a 5.46\% Rank-1 accuracy and 5.97\% mAP, which is an unquestionably poor performance. \textbf{(2)} In contrast, appearance-based methods such as GaitSet, GaitPart, and GaitGL, which use 2D silhouettes as input, achieve significantly higher Rank-1 accuracies of 42.31\%, 33.28\%, and 33.78\%, respectively, outperforming GaitGraph by a considerable margin. This demonstrates the potential of 2D silhouettes in learning temporal information. \textbf{(3)} Furthermore, SMPLGait, which utilizes 3D Mesh along with 2D silhouettes, achieves further improvement as the 3D Mesh helps to learn some Cloth-Changing information. Our HybridGait method also outperforms the other state-of-the-art methods, with a 2.52\% boost in Rank-1 accuracy, 2.81\% boost in Rank-5 accuracy, and higher mAP and mINP by 2.16\% and 2.40\%, respectively. These statistics indicate that our HybridGait method exhibits high efficiency and robustness.


\begin{table}[htbp]
\begin{center}
\begin{NiceTabular}{l|cccc}
\arrayrulecolor{tabcolor}
\toprule[1.4pt]
Methods & Rank-1 & Rank-5 & mAP & mINP \\
\midrule
\midrule

GaitGraph & 7.46 & 15.82 & 8.97 & 5.61 \\
GEINet & 12.31& 27.09 & 13.71 & 9.81 \\
GaitSet & 29.61 & 62.64 & 41.41 & 33.28 \\
GaitPart & 36.52 & 54.78 & 36.50 & 28.64 \\
GaitGL & 37.92 & 53.37 & 37.10 & 29.20 \\
SMPLGait & 48.60 & 68.54 & 49.14 & 40.72 \\
\midrule
HybridGait(Ours) & \textbf{51.12} & \textbf{71.35} & \textbf{51.30} & \textbf{43.12} \\

\bottomrule[1.4pt]
\end{NiceTabular}
\end{center}
\caption{Performance comparison of the state-of-the-art methods on CCGait Dataset.}
\label{tab3}
\end{table}

Table \ref{tab2} displays the evaluation results on the Gait3D dataset. We observe that the model-based method GaitGraph achieves a low accuracy, as it inevitably loses some useful gait information. The appearance-based methods, such as GaitSet, GaitPart, and GaitGL, which use 2D silhouettes as input, obtain better results.
The recent SMPLGait, MTSGait and GaitGCI achieved further improvement due to their specific designed model. Our HybridGait method outperforms the base model SMPLGait, improving Rank-1 by 7.0\% and mAP by 6.13\%. This demonstrates that our method has an overall performance advantage in extracting and combining multi-modality (appearance and model information) for gait recognition.
Note that we have not evaluated the MTSGait and GaitGCI methods on the CCGait dataset since their code is not publicly available. 

In conclusion, our study provides a comprehensive comparison of SOTA gait recognition models. Our findings can serve as a guide for future research aimed at developing more robust models capable of handling variations in view angle and clothing changes over time.


\begin{table}[htbp]
\begin{center}
\begin{NiceTabular}{l|cccc}
\arrayrulecolor{tabcolor}
\toprule[1.4pt]
Methods & Rank-1 & Rank-5 & mAP & mINP \\
\midrule
\midrule
GaitGraph & 6.25 & 16.23 & 5.18 & 2.42 \\
GEINet  & 5.40 & 14.20 & 5.06 & 3.14 \\
\midrule
GaitSet & 36.70 & 58.30 & 30.01 & 17.30 \\
GaitPart & 28.20 & 47.60 & 21.58 & 12.36 \\
GaitGL  & 29.70 & 48.50 & 22.29 & 13.26 \\
CSTL  & 11.70 & 19.20 & 5.59 & 2.59 \\
SMPLGait  & 46.30 & 64.50 & 37.16 & 22.23 \\
MTSGait & 48.70 & 67.10 & 37.63 & 21.92 \\
GaitGCI  & 50.30 & 68.50 & 39.50 & 24.30 \\
\midrule
HybridGait(Ours)& \textbf{53.30} & \textbf{72.00} & \textbf{43.29} & \textbf{26.65} \\

\bottomrule[1.4pt]
\end{NiceTabular}
\end{center}
\caption{Performance comparison of the state-of-the-art methods on Gait3D Dataset.}
\label{tab2}
\end{table}


\subsection{Ablation Study}
To further evaluate the different components in the model, we conduct ablation studies on both Gait3D and our CCGait dataset. The findings, as shown in Table \ref{tab5}, confirm that the HybridGait approach is more effective in overcoming the challenges associated with gait recognition in real-world settings.

\noindent {\bf Effect of Temporal Branch.}
Table \ref{tab4} illustrates that using only the basic appearance branch (Appr) resulted in a Rank-1 accuracy of 42.90\% and mAP of 35.19\% on the Gait3D dataset. However, fusing the basic appearance branch with the temporal branch, which utilizes a Spatial-Temporal Transformer (STT), improved the Rank-1 accuracy to 48.90\% and increased the mAP to 39.90\%. Subsequently, replacing the STT with our proposed  Canonical Alignment Spatial-Temporal Transformer (CA-STT) resulted in further accuracy improvements as a 50.26\% Rank-1 accuracy and 41.36\% mAP. Furthermore, when combined with the appearance branch, the CA-STT achieved a Rank-1 accuracy of 50.28\% with 50.41\% mAP on our CCGait dataset, demonstrating the Temporal branch's ability to effectively capture temporal information from gait data.

\begin{table}[htbp]
\begin{center}
\begin{NiceTabular}{l|c c|c c}
\arrayrulecolor{tabcolor}
\toprule[1.4pt]
\multirow{2}{*}{Methods} & 
\multicolumn{2}{c|}{Gait3D} & \multicolumn{2}{c}{CCGait}\\
\cline{2-5}
& R-1 & mAP &R-1 &mAP\\
\midrule
\midrule
Appr & 42.90 & 35.19 & 44.31 & 36.73\\
Appr + STT & 48.90 & 39.90 & 47.36 & 50.08 \\
Appr + CA-STT & \textbf{50.26} & \textbf{41.36} & \textbf{50.28} & \textbf{50.41} \\

\bottomrule[1.4pt]
\end{NiceTabular}
\end{center}
\caption{Effect of the Temporal Branch with 3D Dynamics.}
\label{tab4}

\end{table}

\noindent {\bf Effect of Projection Branch.}
As presented in Table \ref{tab5}, the inclusion of Projection Branch (SilD) into the basic appearance branch (Appr) leads to better performance, thus validating the effectiveness of SilD in learning more informative appearance features. Furthermore, the combination of SilD with our temporal branch results in a noticeable improvement in performance. Specifically, the projection branch with front view ($0^{\circ}$) achieves a 3.41\% boost in Rank-1 accuracy on Gait3D dataset and a 2.35\% boost on CCGait dataset. Notably, when SilD is added to our proposed CA-STT, our method achieves state-of-the-art performance, highlighting the advantage of our method in gait recognition under challenging conditions.

\begin{table}[htbp]
\begin{center}
\resizebox{\columnwidth}{!}{
\begin{NiceTabular}{l| c c|c c}
\arrayrulecolor{tabcolor}
\toprule[1.4pt]
\multirow{2}{*}{Methods} & 

\multicolumn{2}{c|}{Gait3D} & \multicolumn{2}{c}{CCGait}\\
\cline{2-5}

& R-1 & mAP &R-1 &mAP\\
\midrule
\midrule
Appr & 42.90 & 35.19 & 44.31 & 36.73\\

Appr + SilD & 46.31 & 37.25 & 46.66 & 39.18 \\
Appr + STT + SilD & 52.10 & 41.98 & 50.35 & 50.84 \\
Appr + CA-STT + SilD (Full)& \textbf{53.30} & \textbf{43.29} & \textbf{51.12} & \textbf{51.30}\\

\bottomrule[1.4pt]
\end{NiceTabular}
}
\end{center}
\caption{Effect of the Projection Branch.}
\label{tab5}

\end{table}

\section{Conclusion}
In this paper, we propose the first in-the-wild cloth-changing gait recognition dataset, named CCGait, to facilitate the research community including diverse appearance changes over space and temporal, and other in-the-wild challenges, such as occlusion, in door \& out door, lighting variations. To address the challenges posed by contextual perturbations, we further propose a hybrid approach that fully utilizes a 3D human mesh in both the Temporal branch and the Projection branch. Extensive experiments validate the efficacy of our method on a wide range of benchmarks.

\section{Acknowledgments}
This work was supported by NSFC (No.62303319), Shanghai Sailing Program (21YF1429400,
22YF1428800), Shanghai Local College Capacity Building Program (23010503100), Shanghai
Frontiers Science Center of Human-centered Artificial Intelligence (ShangHAI), MoE Key Laboratory of Intelligent Perception and Human-Machine Collaboration (ShanghaiTech University), and
Shanghai Engineering Research Center of Intelligent Vision and Imaging.

\bibliography{aaai24}

\begin{thebibliography}{62}
\providecommand{\natexlab}[1]{#1}

\bibitem[{An et~al.(2020)An, Yu, Makihara, Wu, Xu, Yu, Liao, and Yagi}]{OU-MVLP-Pose}
An, W.; Yu, S.; Makihara, Y.; Wu, X.; Xu, C.; Yu, Y.; Liao, R.; and Yagi, Y. 2020.
\newblock Performance evaluation of model-based gait on multi-view very large population database with pose sequences.
\newblock \emph{IEEE transactions on biometrics, behavior, and identity science}, 2(4): 421--430.

\bibitem[{Ariyanto and Nixon(2011)}]{model-base}
Ariyanto, G.; and Nixon, M.~S. 2011.
\newblock Model-based 3D gait biometrics.
\newblock In \emph{2011 international joint conference on biometrics (IJCB)}, 1--7. IEEE.

\bibitem[{Arnab et~al.(2021)Arnab, Dehghani, Heigold, Sun, Lu{\v{c}}i{\'c}, and Schmid}]{VIVIT}
Arnab, A.; Dehghani, M.; Heigold, G.; Sun, C.; Lu{\v{c}}i{\'c}, M.; and Schmid, C. 2021.
\newblock Vivit: A video vision transformer.
\newblock In \emph{Proceedings of the IEEE/CVF international conference on computer vision}, 6836--6846.

\bibitem[{Chao et~al.(2019)Chao, He, Zhang, and Feng}]{GaitSet}
Chao, H.; He, Y.; Zhang, J.; and Feng, J. 2019.
\newblock Gaitset: Regarding gait as a set for cross-view gait recognition.
\newblock In \emph{Proceedings of the AAAI conference on artificial intelligence}, volume~33, 8126--8133.

\bibitem[{Clancey(1979)}]{c:79}
Clancey, W.~J. 1979.
\newblock \emph{{Transfer of Rule-Based Expertise through a Tutorial Dialogue}}.
\newblock {Ph.D.} diss., Dept.\ of Computer Science, Stanford Univ., Stanford, Calif.

\bibitem[{Clancey(1983)}]{c:83}
Clancey, W.~J. 1983.
\newblock {Communication, Simulation, and Intelligent Agents: Implications of Personal Intelligent Machines for Medical Education}.
\newblock In \emph{Proceedings of the Eighth International Joint Conference on Artificial Intelligence {(IJCAI-83)}}, 556--560. Menlo Park, Calif: {IJCAI Organization}.

\bibitem[{Clancey(1984)}]{c:84}
Clancey, W.~J. 1984.
\newblock {Classification Problem Solving}.
\newblock In \emph{Proceedings of the Fourth National Conference on Artificial Intelligence}, 45--54. Menlo Park, Calif.: AAAI Press.

\bibitem[{Clancey(2021)}]{c:21}
Clancey, W.~J. 2021.
\newblock {The Engineering of Qualitative Models}.
\newblock Forthcoming.

\bibitem[{Dai et~al.(2017)Dai, Qi, Xiong, Li, Zhang, Hu, and Wei}]{DefConv}
Dai, J.; Qi, H.; Xiong, Y.; Li, Y.; Zhang, G.; Hu, H.; and Wei, Y. 2017.
\newblock Deformable convolutional networks.
\newblock In \emph{Proceedings of the IEEE international conference on computer vision}, 764--773.

\bibitem[{Dou et~al.(2023)Dou, Zhang, Su, Yu, Lin, and Li}]{GaitGCI}
Dou, H.; Zhang, P.; Su, W.; Yu, Y.; Lin, Y.; and Li, X. 2023.
\newblock Gaitgci: Generative counterfactual intervention for gait recognition.
\newblock In \emph{Proceedings of the IEEE/CVF Conference on Computer Vision and Pattern Recognition}, 5578--5588.

\bibitem[{Engelmore and Morgan(1986)}]{em:86}
Engelmore, R.; and Morgan, A., eds. 1986.
\newblock \emph{Blackboard Systems}.
\newblock Reading, Mass.: Addison-Wesley.

\bibitem[{Fan et~al.(2023)Fan, Liang, Shen, Hou, Huang, and Yu}]{OpenGait}
Fan, C.; Liang, J.; Shen, C.; Hou, S.; Huang, Y.; and Yu, S. 2023.
\newblock OpenGait: Revisiting Gait Recognition Towards Better Practicality.
\newblock In \emph{Proceedings of the IEEE/CVF Conference on Computer Vision and Pattern Recognition}, 9707--9716.

\bibitem[{Fan et~al.(2020)Fan, Peng, Cao, Liu, Hou, Chi, Huang, Li, and He}]{GaitPart}
Fan, C.; Peng, Y.; Cao, C.; Liu, X.; Hou, S.; Chi, J.; Huang, Y.; Li, Q.; and He, Z. 2020.
\newblock Gaitpart: Temporal part-based model for gait recognition.
\newblock In \emph{Proceedings of the IEEE/CVF conference on computer vision and pattern recognition}, 14225--14233.

\bibitem[{Gu et~al.(2022)Gu, Chang, Ma, Bai, Shan, and Chen}]{CCVID}
Gu, X.; Chang, H.; Ma, B.; Bai, S.; Shan, S.; and Chen, X. 2022.
\newblock Clothes-changing person re-identification with rgb modality only.
\newblock In \emph{Proceedings of the IEEE/CVF Conference on Computer Vision and Pattern Recognition}, 1060--1069.

\bibitem[{Han et~al.(2022)Han, Cong, Xu, Wang, Yu, and Ma}]{han2022licamgait}
Han, X.; Cong, P.; Xu, L.; Wang, J.; Yu, J.; and Ma, Y. 2022.
\newblock LiCamGait: Gait Recognition in the Wild by Using LiDAR and Camera Multi-modal Visual Sensors.
\newblock arXiv:2211.12371.

\bibitem[{Hasling, Clancey, and Rennels(1984)}]{hcr:83}
Hasling, D.~W.; Clancey, W.~J.; and Rennels, G. 1984.
\newblock Strategic explanations for a diagnostic consultation system.
\newblock \emph{International Journal of Man-Machine Studies}, 20(1): 3--19.

\bibitem[{Hasling et~al.(1983)Hasling, Clancey, Rennels, and Test}]{hcrt:83}
Hasling, D.~W.; Clancey, W.~J.; Rennels, G.~R.; and Test, T. 1983.
\newblock {Strategic Explanations in Consultation---Duplicate}.
\newblock \emph{The International Journal of Man-Machine Studies}, 20(1): 3--19.

\bibitem[{He et~al.(2021)He, Xu, Saito, Soatto, and Tung}]{arch++}
He, T.; Xu, Y.; Saito, S.; Soatto, S.; and Tung, T. 2021.
\newblock Arch++: Animation-ready clothed human reconstruction revisited.
\newblock In \emph{Proceedings of the IEEE/CVF international conference on computer vision}, 11046--11056.

\bibitem[{Hossain et~al.(2010)Hossain, Makihara, Wang, and Yagi}]{OU-ISIR-Cloth}
Hossain, M.~A.; Makihara, Y.; Wang, J.; and Yagi, Y. 2010.
\newblock Clothing-invariant gait identification using part-based clothing categorization and adaptive weight control.
\newblock \emph{Pattern Recognition}, 43(6): 2281--2291.

\bibitem[{Hou et~al.(2020)Hou, Cao, Liu, and Huang}]{GLN}
Hou, S.; Cao, C.; Liu, X.; and Huang, Y. 2020.
\newblock Gait lateral network: Learning discriminative and compact representations for gait recognition.
\newblock In \emph{European conference on computer vision}, 382--398. Springer.

\bibitem[{Huang et~al.(2021{\natexlab{a}})Huang, Zhu, Wang, Wang, Yang, He, Liu, and Feng}]{CSTL}
Huang, X.; Zhu, D.; Wang, H.; Wang, X.; Yang, B.; He, B.; Liu, W.; and Feng, B. 2021{\natexlab{a}}.
\newblock Context-sensitive temporal feature learning for gait recognition.
\newblock In \emph{Proceedings of the IEEE/CVF International Conference on Computer Vision}, 12909--12918.

\bibitem[{Huang et~al.(2021{\natexlab{b}})Huang, Xue, Shen, Tian, Li, Huang, and Hua}]{3DLocal}
Huang, Z.; Xue, D.; Shen, X.; Tian, X.; Li, H.; Huang, J.; and Hua, X.-S. 2021{\natexlab{b}}.
\newblock 3D local convolutional neural networks for gait recognition.
\newblock In \emph{Proceedings of the IEEE/CVF International Conference on Computer Vision}, 14920--14929.

\bibitem[{Iwama et~al.(2012)Iwama, Okumura, Makihara, and Yagi}]{OU-ISLR-LP}
Iwama, H.; Okumura, M.; Makihara, Y.; and Yagi, Y. 2012.
\newblock The OU-ISIR gait database comprising the large population dataset and performance evaluation of gait recognition.
\newblock \emph{IEEE Transactions on Information Forensics and Security}, 7(5): 1511--1521.

\bibitem[{Kingma and Ba(2014)}]{Adam}
Kingma, D.~P.; and Ba, J. 2014.
\newblock Adam: A method for stochastic optimization.
\newblock \emph{arXiv preprint arXiv:1412.6980}.

\bibitem[{Li et~al.(2023{\natexlab{a}})Li, Bian, Liu, Tang, Wang, and Lu}]{li2023niki}
Li, J.; Bian, S.; Liu, Q.; Tang, J.; Wang, F.; and Lu, C. 2023{\natexlab{a}}.
\newblock {NIKI}: Neural Inverse Kinematics with Invertible Neural Networks for 3D Human Pose and Shape Estimation.
\newblock In \emph{Proceedings of the IEEE/CVF Conference on Computer Vision and Pattern Recognition}.

\bibitem[{Li et~al.(2023{\natexlab{b}})Li, Hou, Zhang, Cao, Liu, Huang, and Zhao}]{CCPG}
Li, W.; Hou, S.; Zhang, C.; Cao, C.; Liu, X.; Huang, Y.; and Zhao, Y. 2023{\natexlab{b}}.
\newblock An In-Depth Exploration of Person Re-Identification and Gait Recognition in Cloth-Changing Conditions.
\newblock In \emph{Proceedings of the IEEE/CVF Conference on Computer Vision and Pattern Recognition}, 13824--13833.

\bibitem[{Li et~al.(2022)Li, Makihara, Xu, and Yagi}]{OUMVLP-Mesh}
Li, X.; Makihara, Y.; Xu, C.; and Yagi, Y. 2022.
\newblock Multi-View Large Population Gait Database With Human Meshes and Its Performance Evaluation.
\newblock \emph{IEEE Transactions on Biometrics, Behavior, and Identity Science}, 4(2): 234--248.

\bibitem[{Liao et~al.(2020)Liao, Yu, An, and Huang}]{model}
Liao, R.; Yu, S.; An, W.; and Huang, Y. 2020.
\newblock A model-based gait recognition method with body pose and human prior knowledge.
\newblock \emph{Pattern Recognition}, 98: 107069.

\bibitem[{Lin, Zhang, and Bao(2020)}]{MT3D}
Lin, B.; Zhang, S.; and Bao, F. 2020.
\newblock Gait recognition with multiple-temporal-scale 3d convolutional neural network.
\newblock In \emph{Proceedings of the 28th ACM international conference on multimedia}, 3054--3062.

\bibitem[{Lin, Zhang, and Yu(2021)}]{GaitGL}
Lin, B.; Zhang, S.; and Yu, X. 2021.
\newblock Gait recognition via effective global-local feature representation and local temporal aggregation.
\newblock In \emph{Proceedings of the IEEE/CVF International Conference on Computer Vision}, 14648--14656.

\bibitem[{Loper et~al.(2023)Loper, Mahmood, Romero, Pons-Moll, and Black}]{smpl}
Loper, M.; Mahmood, N.; Romero, J.; Pons-Moll, G.; and Black, M.~J. 2023.
\newblock SMPL: A skinned multi-person linear model.
\newblock In \emph{Seminal Graphics Papers: Pushing the Boundaries, Volume 2}, 851--866.

\bibitem[{Makihara, Mannami, and Yagi(2011)}]{OU-ISIR-MV}
Makihara, Y.; Mannami, H.; and Yagi, Y. 2011.
\newblock Gait analysis of gender and age using a large-scale multi-view gait database.
\newblock In \emph{Computer Vision--ACCV 2010: 10th Asian Conference on Computer Vision, Queenstown, New Zealand, November 8-12, 2010, Revised Selected Papers, Part II 10}, 440--451. Springer.

\bibitem[{Moon, Chang, and Lee(2019{\natexlab{a}})}]{PoseNet}
Moon, G.; Chang, J.~Y.; and Lee, K.~M. 2019{\natexlab{a}}.
\newblock Camera distance-aware top-down approach for 3d multi-person pose estimation from a single rgb image.
\newblock In \emph{Proceedings of the IEEE/CVF international conference on computer vision}, 10133--10142.

\bibitem[{Moon, Chang, and Lee(2019{\natexlab{b}})}]{mmhuman3d}
Moon, G.; Chang, J.~Y.; and Lee, K.~M. 2019{\natexlab{b}}.
\newblock Camera distance-aware top-down approach for 3d multi-person pose estimation from a single rgb image.
\newblock In \emph{Proceedings of the IEEE/CVF international conference on computer vision}, 10133--10142.

\bibitem[{{NASA}(2015)}]{c:23}
{NASA}. 2015.
\newblock Pluto: The 'Other' Red Planet.
\newblock \url{https://www.nasa.gov/nh/pluto-the-other-red-planet}.
\newblock Accessed: 2018-12-06.

\bibitem[{Rice(1986)}]{r:86}
Rice, J. 1986.
\newblock {Poligon: A System for Parallel Problem Solving}.
\newblock Technical Report KSL-86-19, Dept.\ of Computer Science, Stanford Univ.

\bibitem[{Robinson(1980{\natexlab{a}})}]{r:80}
Robinson, A.~L. 1980{\natexlab{a}}.
\newblock New Ways to Make Microcircuits Smaller.
\newblock \emph{Science}, 208(4447): 1019--1022.

\bibitem[{Robinson(1980{\natexlab{b}})}]{r:80x}
Robinson, A.~L. 1980{\natexlab{b}}.
\newblock {New Ways to Make Microcircuits Smaller---Duplicate Entry}.
\newblock \emph{Science}, 208: 1019--1026.

\bibitem[{Sarkar et~al.(2005)Sarkar, Phillips, Liu, Vega, Grother, and Bowyer}]{USFHumanID}
Sarkar, S.; Phillips, P.~J.; Liu, Z.; Vega, I.~R.; Grother, P.; and Bowyer, K.~W. 2005.
\newblock The humanid gait challenge problem: Data sets, performance, and analysis.
\newblock \emph{IEEE transactions on pattern analysis and machine intelligence}, 27(2): 162--177.

\bibitem[{Shiraga et~al.(2016)Shiraga, Makihara, Muramatsu, Echigo, and Yagi}]{GEINet}
Shiraga, K.; Makihara, Y.; Muramatsu, D.; Echigo, T.; and Yagi, Y. 2016.
\newblock Geinet: View-invariant gait recognition using a convolutional neural network.
\newblock In \emph{2016 international conference on biometrics (ICB)}, 1--8. IEEE.

\bibitem[{Sun et~al.(2021)Sun, Bao, Liu, Fu, Black, and Mei}]{ROMP}
Sun, Y.; Bao, Q.; Liu, W.; Fu, Y.; Black, M.~J.; and Mei, T. 2021.
\newblock Monocular, one-stage, regression of multiple 3d people.
\newblock In \emph{Proceedings of the IEEE/CVF international conference on computer vision}, 11179--11188.

\bibitem[{Takemura et~al.(2018)Takemura, Makihara, Muramatsu, Echigo, and Yagi}]{OU-MVLP}
Takemura, N.; Makihara, Y.; Muramatsu, D.; Echigo, T.; and Yagi, Y. 2018.
\newblock Multi-view large population gait dataset and its performance evaluation for cross-view gait recognition.
\newblock \emph{IPSJ transactions on Computer Vision and Applications}, 10: 1--14.

\bibitem[{Tan et~al.(2006)Tan, Huang, Yu, and Tan}]{CASIA-C}
Tan, D.; Huang, K.; Yu, S.; and Tan, T. 2006.
\newblock Efficient night gait recognition based on template matching.
\newblock In \emph{18th international conference on pattern recognition (ICPR'06)}, volume~3, 1000--1003. IEEE.

\bibitem[{Teepe et~al.(2021)Teepe, Khan, Gilg, Herzog, H{\"o}rmann, and Rigoll}]{GaitGraph}
Teepe, T.; Khan, A.; Gilg, J.; Herzog, F.; H{\"o}rmann, S.; and Rigoll, G. 2021.
\newblock Gaitgraph: Graph convolutional network for skeleton-based gait recognition.
\newblock In \emph{2021 IEEE International Conference on Image Processing (ICIP)}, 2314--2318. IEEE.

\bibitem[{Tsuji, Makihara, and Yagi(2010)}]{OU-ISIR-Speed}
Tsuji, A.; Makihara, Y.; and Yagi, Y. 2010.
\newblock Silhouette transformation based on walking speed for gait identification.
\newblock In \emph{2010 IEEE Computer Society Conference on Computer Vision and Pattern Recognition}, 717--722. IEEE.

\bibitem[{Uddin et~al.(2018)Uddin, Ngo, Makihara, Takemura, Li, Muramatsu, and Yagi}]{OU-LP-Bag}
Uddin, M.~Z.; Ngo, T.~T.; Makihara, Y.; Takemura, N.; Li, X.; Muramatsu, D.; and Yagi, Y. 2018.
\newblock The ou-isir large population gait database with real-life carried object and its performance evaluation.
\newblock \emph{IPSJ Transactions on Computer Vision and Applications}, 10(1): 1--11.

\bibitem[{Vaswani et~al.(2017)Vaswani, Shazeer, Parmar, Uszkoreit, Jones, Gomez, Kaiser, and Polosukhin}]{c:22}
Vaswani, A.; Shazeer, N.; Parmar, N.; Uszkoreit, J.; Jones, L.; Gomez, A.~N.; Kaiser, L.; and Polosukhin, I. 2017.
\newblock Attention Is All You Need.
\newblock arXiv:1706.03762.

\bibitem[{Wang et~al.(2020)Wang, Sun, Cheng, Jiang, Deng, Zhao, Liu, Mu, Tan, Wang et~al.}]{HRNet-Segmentation}
Wang, J.; Sun, K.; Cheng, T.; Jiang, B.; Deng, C.; Zhao, Y.; Liu, D.; Mu, Y.; Tan, M.; Wang, X.; et~al. 2020.
\newblock Deep high-resolution representation learning for visual recognition.
\newblock \emph{IEEE transactions on pattern analysis and machine intelligence}, 43(10): 3349--3364.

\bibitem[{Wang et~al.(2003)Wang, Tan, Ning, and Hu}]{CASIA-A}
Wang, L.; Tan, T.; Ning, H.; and Hu, W. 2003.
\newblock Silhouette analysis-based gait recognition for human identification.
\newblock \emph{IEEE transactions on pattern analysis and machine intelligence}, 25(12): 1505--1518.

\bibitem[{Xu and Zhu(2021)}]{DeepChange}
Xu, P.; and Zhu, X. 2021.
\newblock Deepchange: A large long-term person re-identification benchmark with clothes change.
\newblock \emph{arXiv preprint arXiv:2105.14685}.

\bibitem[{Yam, Nixon, and Carter(2004)}]{AutoMated}
Yam, C.; Nixon, M.~S.; and Carter, J.~N. 2004.
\newblock Automated person recognition by walking and running via model-based approaches.
\newblock \emph{Pattern recognition}, 37(5): 1057--1072.

\bibitem[{Yamauchi, Bhanu, and Saito(2009)}]{Recognition}
Yamauchi, K.; Bhanu, B.; and Saito, H. 2009.
\newblock Recognition of walking humans in 3D: Initial results.
\newblock In \emph{2009 IEEE Computer Society Conference on Computer Vision and Pattern Recognition Workshops}, 45--52. IEEE.

\bibitem[{Yu, Tan, and Tan(2006)}]{CASIA-B}
Yu, S.; Tan, D.; and Tan, T. 2006.
\newblock A framework for evaluating the effect of view angle, clothing and carrying condition on gait recognition.
\newblock In \emph{18th international conference on pattern recognition (ICPR'06)}, volume~4, 441--444. IEEE.

\bibitem[{Zhang et~al.(2023)Zhang, Shi, Ma, Xu, Yu, and Wang}]{zhang2023ikol}
Zhang, J.; Shi, Y.; Ma, Y.; Xu, L.; Yu, J.; and Wang, J. 2023.
\newblock IKOL: Inverse kinematics optimization layer for 3D human pose and shape estimation via Gauss-Newton differentiation.
\newblock In \emph{Proceedings of the AAAI Conference on Artificial Intelligence (AAAI)}.

\bibitem[{Zhang et~al.(2021)Zhang, Wang, Wang, Zeng, and Liu}]{FairMOT}
Zhang, Y.; Wang, C.; Wang, X.; Zeng, W.; and Liu, W. 2021.
\newblock Fairmot: On the fairness of detection and re-identification in multiple object tracking.
\newblock \emph{International Journal of Computer Vision}, 129: 3069--3087.

\bibitem[{Zhang et~al.(2019)Zhang, Tran, Yin, Atoum, Liu, Wan, and Wang}]{FVG}
Zhang, Z.; Tran, L.; Yin, X.; Atoum, Y.; Liu, X.; Wan, J.; and Wang, N. 2019.
\newblock Gait recognition via disentangled representation learning.
\newblock In \emph{Proceedings of the IEEE/CVF conference on computer vision and pattern recognition}, 4710--4719.

\bibitem[{Zheng et~al.(2021)Zheng, Zhu, Mendieta, Yang, Chen, and Ding}]{3dhuman}
Zheng, C.; Zhu, S.; Mendieta, M.; Yang, T.; Chen, C.; and Ding, Z. 2021.
\newblock 3d human pose estimation with spatial and temporal transformers.
\newblock In \emph{Proceedings of the IEEE/CVF International Conference on Computer Vision}, 11656--11665.

\bibitem[{Zheng et~al.(2022{\natexlab{a}})Zheng, Liu, Gu, Sun, Gan, Zhang, Liu, and Yan}]{MTSGait}
Zheng, J.; Liu, X.; Gu, X.; Sun, Y.; Gan, C.; Zhang, J.; Liu, W.; and Yan, C. 2022{\natexlab{a}}.
\newblock Gait recognition in the wild with multi-hop temporal switch.
\newblock In \emph{Proceedings of the 30th ACM International Conference on Multimedia}, 6136--6145.

\bibitem[{Zheng et~al.(2022{\natexlab{b}})Zheng, Liu, Liu, He, Yan, and Mei}]{Gait3D}
Zheng, J.; Liu, X.; Liu, W.; He, L.; Yan, C.; and Mei, T. 2022{\natexlab{b}}.
\newblock Gait recognition in the wild with dense 3d representations and a benchmark.
\newblock In \emph{Proceedings of the IEEE/CVF Conference on Computer Vision and Pattern Recognition}, 20228--20237.

\bibitem[{Zhu, Zheng, and Nevatia(2023)}]{HBS}
Zhu, H.; Zheng, Z.; and Nevatia, R. 2023.
\newblock Gait recognition using 3-d human body shape inference.
\newblock In \emph{Proceedings of the IEEE/CVF Winter Conference on Applications of Computer Vision}, 909--918.

\bibitem[{Zhu et~al.(2021{\natexlab{a}})Zhu, Guo, Yang, Huang, Deng, Huang, Du, Lu, and Zhou}]{GREW}
Zhu, Z.; Guo, X.; Yang, T.; Huang, J.; Deng, J.; Huang, G.; Du, D.; Lu, J.; and Zhou, J. 2021{\natexlab{a}}.
\newblock Gait recognition in the wild: A benchmark.
\newblock In \emph{Proceedings of the IEEE/CVF international conference on computer vision}, 14789--14799.

\bibitem[{Zhu et~al.(2021{\natexlab{b}})Zhu, Guo, Yang, Huang, Deng, Huang, Du, Lu, and Zhou}]{GRAW}
Zhu, Z.; Guo, X.; Yang, T.; Huang, J.; Deng, J.; Huang, G.; Du, D.; Lu, J.; and Zhou, J. 2021{\natexlab{b}}.
\newblock Gait recognition in the wild: A benchmark.
\newblock In \emph{Proceedings of the IEEE/CVF international conference on computer vision}, 14789--14799.

\end{thebibliography}

\end{document}